\pdfoutput=1
\documentclass{isprs}
\usepackage{subfigure}
\usepackage{setspace}
\usepackage{geometry} 
\usepackage{epstopdf}
\usepackage[labelsep=period]{caption}  
\usepackage{amssymb}
\usepackage{booktabs}
\usepackage{multirow}
\usepackage{amsmath}
\usepackage{lscape}
\usepackage{rotating}
\usepackage[english]{babel}
\usepackage{hyperref}

\DeclareMathOperator*{\argmin}{arg\,min}

\usepackage{diagbox}
\usepackage{xcolor}

\newcommand{\bp}{\textbf{p}}
\newcommand{\bx}{\textbf{x}}
\newcommand{\be}{\textbf{e}}

\newcommand\crule[3][black]{\textcolor[HTML]{#1}{\rule{#2}{#3}}}

\usepackage{pifont}
\newcommand{\cmark}{\checkmark} 
\newcommand{\xmark}{\ding{55}}%

\newcommand{\datasetsurl}{\url{https://pix4d.com/research}}

\begin{document}

\title{Classification of Aerial Photogrammetric 3D Point Clouds}
	
\author{
C. Becker \textsuperscript{a}, N. H{\"a}ni \textsuperscript{b}\thanks{This work was done while the author was working 
at Pix4D SA}\ , E. Rosinskaya 
\textsuperscript{a}, E. d'Angelo 
\textsuperscript{a}, C. Strecha \textsuperscript{a}}

\address
{
	\textsuperscript{a }Pix4D SA, EPFL Innovation Park, Building F, 1015 Lausanne, 
	Switzerland \\
	\textsuperscript{b} University of Minnesota\\
	(carlos.becker, elena.rosinskaya, emmanuel.dangelo, 
	christoph.strecha)@pix4d.com
	\\
	haeni001@umn.edu\\
}

\commission{II }{II} 
\workinggroup{II/2} 
\icwg{}   

\abstract
{
We present a powerful method to extract per-point semantic class labels from aerial 
photogrammetry data. Labelling this kind of
data is important for tasks such as environmental modelling, object classification and scene 
understanding. 
Unlike previous point cloud classification methods that rely exclusively on geometric 
features, we show that 
incorporating color information yields a significant increase in accuracy in detecting 
semantic classes.
We test our classification method on three real-world photogrammetry datasets 
that were generated with Pix4Dmapper Pro, and  
with varying point densities.
We show that off-the-shelf machine learning techniques coupled with our new features allow 
us to train highly accurate classifiers that generalize well to unseen data, processing point clouds containing 10 
million points  in less than 3 minutes on a desktop computer.
}


\maketitle

\section{Introduction}
\label{sec:intro}

Extraction of semantic information from point clouds enables us to understand a scene, classify objects and generate high-level models with CAD-like geometries from them.
It can also provide a significant improvement to existing algorithms, such as those used to construct Digital Terrain 
Models (DTMs) from Digital Surface Models (DSMs)~\cite{hutchison_variational_2009}.
With the growing popularity of laser scanners, the availability of drones as surveying tools, and the rise of 
commercial photogrammetry software capable of generating millions of points from images, there exists an increasing 
need for fully automated extraction of semantic information from this kind of data.
Although some of the commercial photogrammetry software available today offer tools such as automated DTM 
extraction~\cite{Pix4D,Agisoft}, semantic classification is typically left to specialized software 
packages~\cite{eCognition,GlobalMapper} that rely on $2.5$D orthomosaics and DSMs as an input.

The need for semantic modeling of 3D point data has inspired many research and application engineers to model 
specific structures. Often the proposed solutions were handcrafted to the application at hand: buildings have been 
modeled by using common image processing techniques such as edge detection~\cite{haala_3d_1998,Brenner00towardsfully} 
or by fitting planes to point clouds~\cite{rusu_towards_2007}; road networks have been modeled by handcrafted features 
and DTM algorithms used heuristics about the size of objects to create a DTM from a DSM. While 
successful and valuable, these approaches are inherently limited since they cannot be easily applied to detect 
new classes of objects. The huge boost in the performance of machine learning algorithms over the last years allows 
for more flexible and general learning and classification algorithms. If supervised or semi-supervised 
learning 
and especially classification becomes fast and reliable, machine learning approaches to point cloud classification will 
find 
their way into common photogrammetric workflows. Therefore we focus here on machine learning techniques, that will allow the users to detect objects categories of their own choice.

In this paper we present a method to classify aerial photogrammetry point clouds. Our approach exploits both geometric 
and color information to classify individual points as belonging to one of the following classes extracted from the LAS 
standard: \emph{buildings, terrain, high vegetation, roads, human made objects} or \emph{cars}. Unlike previous point 
cloud classification methods that rely exclusively on geometric features,  we show that incorporating color information 
yields a significant increase in accuracy.

We evaluate our approach on three challenging datasets and show that off-the-shelf machine learning techniques together 
with our new features result in highly accurate and efficient classifiers that generalize well to unseen data. 
The datasets used for evaluation are publicly available at \datasetsurl.

\section{Related work}
\label{sec:relwork}

Methods used to extract semantic information from point clouds can be split into two groups: those that try to segment coherent objects from a scene, and those that focus on assigning an individual class label to each point.
Early works using the first approach often converted the point data into a regular 2.5D height grid so that standard 
image processing techniques, e.g., edge detection, can be 
applied~\cite{haala_3d_1998,Haala99extractionof,wang_building_2000}.
A scan line based approach~\cite{sithole_automatic_2003} was proposed for structure detection in urban scenes. Building 
extraction approaches typically use geometric primitives during the segmentation step. A multitude of such primitives 
has been proposed, both in 2D, such as planes and polyhedral~\cite{vosselman_3d_2001,dorninger_3d_2007}, and in
3D~\cite{lafarge_creating_2012,xiao_reconstructing_2014}.
In~\cite{rusu_towards_2007} the authors fit sampled parametric models to the data for object recognition. Similarly, \cite{oesau_object_2016} investigates supervised machine learning techniques to represent small indoor datasets with planar models for object recognition.

The second type of methods assign a label to each point in the point cloud. Typically this is done with supervised 
machine learning techniques, requiring training on labeled data from which a classification model is learned and then 
applied to new, unseen data to predict the label of each point.
Binary classification has been explored in environmental monitoring to extract road 
surfaces~\cite{shu_pairwise-svm_2016}, tree species~\cite{bohm_iqmulus_2016,liu2015classification}, land 
cover~\cite{zhou_land_2016}, and 
construction sites~\cite{xu_classification_2016}.
Several other authors employed a multiclass
setting to classify multiple types of objects and 
structures~\cite{brodu_3d_2012,weinmann_contextual_2015,hackel_fast_2016}, which we adopt in this paper. In particular, 
we follow the work of~\cite{weinmann_feature_2013}, which introduced local geometric features that were used to train a 
Random Forest (RF) classifier for single terrestrial LiDAR scans. Their set of features was extended later
by~\cite{hackel_fast_2016}. Examples of other feature sets used in the point classification context are \textit{Fast 
Point Feature Histogram}  (FPFH)~\cite{rusu_fast_2009} or \textit{Color Signature of Histogram of Orientations} 
(SHOT)~\cite{tombari_unique_2010}.
All these methods use handcrafted features that can be considered suboptimal when compared to more recent deep 
learning techniques~\cite{hu2016deep,qi2016pointnet}, which learn features directly on image or point cloud data. Those 
approaches have not been considered here, since they require large computational power to train the classifier, and may 
be restrictive at prediction time, depending on the hardware available.

The ambiguity of the classification task can be minimized by modeling also the spatial correlations between the 
different class labels. Spatial priors are used in~\cite{shapovalov_cutting-plane_2011} to classify LiDAR data and 
in~\cite{niemeyer_contextual_2014} the authors apply Conditional Random Field (CRF) priors to model different 
probabilities that neighboring labels can have. While those methods show reasonable classification improvements, they 
are computationally expensive and not easy to parallelize.

In this paper we extend the work on geometric features by~\cite{weinmann_feature_2013,hackel_fast_2016} and show that 
incorporating color information provides a significant boost in prediction accuracy, while keeping a low computational 
load at prediction time. In the following sections we describe our method and present the results obtained on three 
photogrammetry datasets.

\section{Method}
\label{sec:method}

Our approach combines geometric and color features that are fed to a classifier to predict 
the class of each point in the point cloud. The geometric features are those introduced in 
\cite{hackel_fast_2016}, which are computed at multiple scales, as explained shortly further. To 
exploit color information, we compute 
additional color features, based on the color of the respective point and its neighbors.

In the next sections we describe the geometric features introduced in \cite{hackel_fast_2016}.
We then show how our color features are computed and discuss implementation details.

\subsection{Geometric Features}
Our approach computes geometric features at different scales to capture details at varying 
spatial resolutions. Below we first describe how features are computed for a single scale, 
and then we show how the scale pyramid is constructed.

We follow the method proposed in \cite{weinmann_feature_2013} and later in 
\cite{hackel_fast_2016}. To compute the features for a point $\bx$, we first obtain its 
neighboring points $\mathcal{S}_{\bx} = \{\bp_1, \bp_2, \dots, \bp_k\}$. This set is used to 
compute a local 3D structure covariance tensor 
\begin{equation}
C_{\bx} = \frac{1}{k} \sum_{i=1}^{k}(\bp_i - \hat{\bp})(\bp_i - \hat{\bp})^T , 
\end{equation}
where
$\hat{\bp} = \argmin_{\bp} \sum_{i=1}^{k}\left| \left| \bp_i - \bp \right| \right| $ is the medoid of 
$\mathcal{S}_{\bx}$.

The eigenvalues $\lambda_1 \geq \lambda_2 \geq \lambda_3 \geq 0$, unit-sum normalized, and 
the corresponding eigenvectors $\be_1, \be_2, \be_3$ of $C_{\bx}$ are used to compute the 
local geometry features shown in Table~\ref{tab:features}.

{\renewcommand{\arraystretch}{1.5}
\begin{table}[tbp]
	\footnotesize
	\begin{tabular}{|c|c|c|}
		\hline 
		\multirow{8}{*}{Covariance}	& Omnivariance & $(\lambda_1 \cdot \lambda_2 \cdot 
		\lambda_3)^\frac{1}{3}$ \\  
		& Eigenentropy & $-\sum_{i=1}^{3} \lambda_i \cdot \ln(\lambda_i)$ \\
		& Anisotropy & $(\lambda_1 - \lambda_3) / \lambda_1$ \\
		& Planarity & $(\lambda_2 - \lambda_3) / \lambda_1$ \\
		& Linearity & $(\lambda_1 - \lambda_2) / \lambda_1$ \\
		& Surface variation & $\lambda_3$ \\
		& Scatter & $\lambda_3 / \lambda_1$ \\
		& Verticality & $1 - \left| \left\langle \left[ 0, 0, 1 \right], \be_3
		\right\rangle \right| $\\
		\hline 
		\multirow{4}{*}{Moments}	& $1^\text{st}$ order, $1^\text{st}$ axis & $\sum_{\bp 
		\in 
		\mathcal{S}_{\bx}} \left\langle \bp - \hat{\bp}, \be_1 \right\rangle$ \\  
		& $1^\text{st}$ order, $2^\text{st}$ axis & $\sum_{\bp \in \mathcal{S}_{\bx}} 
		\left\langle 
		\bp - \hat{\bp}, \be_2 \right\rangle$ \\
		& $2^\text{st}$ order, $1^\text{st}$ axis & $\sum_{\bp \in \mathcal{S}_{\bx}} 
		\left\langle 
		\bp - \hat{\bp}, \be_1 \right\rangle ^2$ \\
		& $2^\text{st}$ order, $2^\text{st}$ axis & $\sum_{\bp \in \mathcal{S}_{\bx}} 
		\left\langle 
		\bp - \hat{\bp}, \be_2 \right\rangle ^2$ \\
		\hline 
		\multirow{3}{*}{Height}	& Vertical Range & $z_{\text{max} \{\mathcal{S}_{\bx} \} } - z_{\text{min}\{\mathcal{S}_{\bx} \}}$ \\  
		& Height below & $z_{\bp} - z_{{\text{min} \{\mathcal{S}_{\bx} \} }} $ \\
		& Height above & $z_{\text{max} \{\mathcal{S}_{\bx} \}} - z_{\bp}$ \\
		\hline
		\hline  
		\multirow{2}{*}{Color}	& Point color & $[H_{\bx}, S_{\bx}, V_{\bx}]$ \\  
		& Neighborhood colors & $ \frac{1}{\left| \mathcal{N}_{\bx}(r) \right|} \sum_{\bp 
		\in 
		\mathcal{N}_{\bx}(r)} [H,S,V]_{\bp}$ \\  
		\hline 
	\end{tabular} 
	
	\caption{Our set of geometric (top) and color features (bottom) computed for points in local neighborhood $\mathcal{S}_{\bx}$.}
	\label{tab:features}
\end{table}
{\renewcommand{\arraystretch}{1}

Besides the features based on the eigenvalues and eigenvectors of $C_{\bx}$, features based
on the $z$ coordinate of the point are used to increase their discriminative power.
We have slightly changed the geometry feature set from \cite{hackel_fast_2016} and removed the sum of eigenvalues 
because it is constant since the eigenvalues are normalized to unit sum.

\paragraph{Multi-scale Pyramid} To incorporate information at different scales we follow 
the multi-scale approach of \cite{hackel_fast_2016}, which has shown to be more 
computationally efficient than that of~\cite{weinmann_distinctive_2015}. Instead of 
computing the geometric features of Table~\ref{tab:features} at a single scale, we 
successively downsample the original point cloud to create a multi-scale pyramid with  
decreasing point densities. The geometric features described earlier are computed at each 
pyramid level and later concatenated.

\paragraph{Pyramid Scale Selection} In order to generalize over different point clouds with varying 
spatial resolution, we need to choose a fixed set of pyramid levels. This is particularly 
important when dealing with data with varying Ground Sampling Distance (GSD), which affects the 
spatial resolution of the point cloud.
The GSD is a characteristic of the images used to generate the pointcloud. It is the distance between two consecutive pixel centers measured in the 
orthographic projection of the images onto the \textit{Digital Surface Model} (DSM). 
Among other factors, the GSD depends on the altitude from which the aerial photos were taken. 
With this in mind, we set the starting resolution of the pyramid to 
four times the largest GSD in our datasets, or $4 \times 5.1~\textrm{cm} = 20.4$ centimeters.
In total we compute $9$ scales, with a downsampling factor of $2$. With these values we were able to capture changes in 
patterns of surfaces and objects which vary with distance (e.g. buildings have significant height variations at the 
scale of dozens meters, while cars, trees do at only a few meters).

\subsection{Color Features}
To increase the discriminative power of the feature set we combine the geometric features 
introduced above with color features. Our color features are computed in the HSV color space 
first introduced by \cite{smith_color_1978}, since the analysis of the Pearson product-moment correlation coefficient 
and 
the Fisher information of our training data showed that we should expect higher information 
gain from the HSV over RGB color space.

Besides the  HSV color values of the point itself, we compute the average color values 
of the neighboring points in the original point cloud (i.e. not downsampled). These points 
are selected as the points within a certain radius around the query point. 
We experimented with the radii $0.4$m, $0.6$m and $0.9$m to balance between speed and accuracy of the classification.

\subsection{Training and Classification}
We use supervised machine learning techniques to train our classifier.
We experiment with two well-known ensemble methods: Random Forest (RF) and Gradient Boosted Trees (GBT) \cite{Hastie01elements}. Though RF has been used extensively in point cloud classification \cite{weinmann_distinctive_2015,hackel_fast_2016}, we 
provide a comparison to GBT and show that the latter can achieve higher accuracies at a similar computational complexity.

Both RF and GBT can generate conditional probabilities and are applicable to multi-class 
classification problems. They are easily parallelized and are available as reusable software 
packages in different programming languages.

Random Forest (RF) \cite{breiman2001random} is a very successful learning method that trains an ensemble of decision 
trees on random subsets of the training data. The output of a RF is the average of the predictions of all the decision 
trees in the ensemble, which has the effect of reducing the overall variance of the classifier.

On the other hand, the Gradient Boosted Trees (GBT) method trains an ensemble of trees by minimizing 
its loss over the training data in a greedy fashion~\cite{Hastie01elements}. GBT has been 
described as one of the best \emph{off-the-shelf} classification methods and it has been shown to 
perform similarly or better than RF in various classification 
tasks~\cite{Caruana06empirical}.

\subsection{Implementation Details}
We implemented our software in C++ to ease its later integration 
into the Pix4DMapper software. For prototyping and evaluation we used 
Julia~\cite{bezanson2014julia}.
For fast neighbor search we used the header-only 
\textit{nanoflann} library~\footnote{https://github.com/jlblancoc/nanoflann)} which implements a kd-tree search 
structure.

The implementation of the RF comes from the \textit{ETH Random Forest Library} 
\footnote{http://www.prs.igp.ethz.ch/research/Source\_code\_and\_datasets.html}.
We parallelized training and prediction, reducing computation times significantly. 
For GBT we used Microsoft's 
\emph{LightGBM}~\footnote{https://github.com/Microsoft/LightGBM}.

\subsection{Experimental Setup}
\label{sec:experimentalsetup}
To evaluate our method  we test different combinations of feature sets and classifiers on photogrammetry data. We compare two different setups to 
evaluate the performance of our approach: within the same dataset, or \emph{intra-dataset}, and across different 
datasets, 
or \emph{inter-dataset}. For training we sampled 10k points of each class at random, resulting in 60k training 
samples.

The different feature sets used in our experiments are summarized below:
\begin{itemize}
	\item \textbf{Geometric features} ($\mathcal{G}$): the geometric eigenvalue-based features shown in 
	Table~\ref{tab:features}. We use $k=10$ neighbors to construct $S_\bx$.
	\item \textbf{Point color} ($\mathcal{C}_p$): HSV color values of the respective 3D point.
	\item \textbf{Point and Neighborhood color} ($\mathcal{C}_{\mathcal{N}(r)}$): $\mathcal{C}_p$ set added with 
	averaged HSV color values of the neighboring 
	points within the radius $r$ around the respective 3D point.
	\item \textbf{All features}: Concatenation of all geometric, point color and neighborhood color features for radii 
	$0.4$m, $0.6$m and $0.9$m.
\end{itemize}

\paragraph{Intra-dataset experiments}
\label{sec:experiments-intratest}
In this setup we divide each dataset into two physically disjoint point clouds. We first find a splitting vertical 
plane such that the resulting point clouds are as similar as possible with respect to the number of points per class. 
More specifically, we solve for the vertical plane $\hat{p}$ 
\begin{equation}
\hat{p} = \argmin_{p \in \mathcal{P}} \Bigg[ 
\max_{c \in \mathcal{Y}} \Big| \frac{1}{\#c} \sum_{\bx_i \in p^+} (y_i = c) \: - \: \frac{1}{2}  \Big| 
\Bigg] \:,
\end{equation}
where $\#c$ is the number of points of class $c$ in the whole point cloud, $\mathcal{Y}$ is the set of all classes 
present in the point cloud, $p^+$ is the set of points falling on one side of the plane $p$, and $\mathcal{P}$ is a set 
of potential vertical planes of different offsets and rotations.

We then train on one of the splits and test on the other. 

\paragraph{Inter-dataset experiments}
\label{sec:experiments-crosstest}
To test the generalization capabilities of our approach to new unseen point clouds we also experiment with a 
leave-one-out evaluation methodology: we train on two point clouds and test on the remaining one.

\section{Results}
\label{sec:results}

In this section we describe first the datasets and classification methods used for the experiments. Next, we show that our implementation is able to reproduce the results presented in~\cite{hackel_fast_2016} on the Paris-rue-Madame dataset. For this dataset we use purely geometric features, as no color information is available.
Finally, we evaluate our approach on challenging aerial photogrammetry point clouds. Our experiments demonstrate that using color information boosts performance significantly, both quantitatively and qualitatively.

\subsection{Datasets}
Table~\ref{tab:datasets} shows the characteristics of the datasets employed for evaluation. The Paris-rue-Madame 
dataset~\cite{serna_paris-rue-madame_2014} does not contain color information and was solely used to verify that our 
geometric features perform as well as those of~\cite{hackel_fast_2016}.

\begin{table}[tbp]
	\centering
	\footnotesize
	\begin{tabular}{lcccc}
		\toprule
		\multirow{2}{*}{Dataset} & \multirow{2}{*}{Acquisition} & 
			\multirow{2}{*}{Color} & \multirow{2}{*}{\# points} & GSD\\
		& & &  & $\big[\frac{cm}{pixel}\big]$\\
		\midrule
		Paris-rue-Madame & Laser scan & no & 20M & N/A\\
		Ankeny & Aerial images & yes & 9.0M & 2.3\\
		Buildings & Aerial images & yes & 3.4M & 1.8\\
		Cadastre & Aerial images & yes & 5.8M & 5.1\\
		\bottomrule
	\end{tabular}
	\caption{Point cloud datasets used for evaluation.}
	\label{tab:datasets}
\end{table}

\begin{table}[tbp]
	\centering
	\footnotesize
	\begin{tabular}{lccc}
		\toprule
		\multirow{1}{*}{Feature} & Ankeny & Buildings & Cadastre\\
		\midrule
		Roads & \cmark & \cmark & \cmark \\
		Ground/Grass on flatland & \cmark & \cmark & \cmark \\
		Ground/Grass on slopes & \xmark & \xmark & \cmark \\
		Gray/white Roofs & \cmark & \cmark & \cmark \\
		Red Roofs & \xmark & \cmark & \cmark \\
		Cropland & \cmark & \xmark & \xmark \\
		\bottomrule
	\end{tabular}
	\caption{Point cloud dataset content break down. The datasets are heterogeneous and contain different objects and types of terrain.}
	\label{tab:datasets-cnt}
\end{table}

Our main interest is the aerial photogrammetry and the three last datasets of Table~\ref{tab:datasets}. 
The images were processed with Pix4Dmapper Pro to obtain their respective dense point clouds that were used as the input for our approach. 
Note that the GSD varies significantly between datasets. A 3D visualization of each of these point clouds is
presented in Fig.~\ref{fig:ankeny-qualit}(a), Fig.~\ref{fig:buildings-qualit}(a), Fig.~\ref{fig:cadastre-qualit}(a).

Moreover, each dataset contains different types of objects and terrain surfaces as shown in 
Table~\ref{tab:datasets-cnt}. For example, while all datasets contain roads, cropland only appears in one of them. This 
table will be useful later to analyze the performance of our approach on each dataset.

We have made the three photogrammetry datasets publicly available at \datasetsurl.

\subsection{Classifier Parameters}
For both GBT and RF we used 100 trees, and at each split half of the features were picked at random as possible 
candidates. For RF the maximum tree depth was set to $30$. For GBT we set the maximum number of leaves to $16$, 
learning rate to $0.2$ and the bagging fraction to $0.5$. These parameters were fixed for all the experiments.

\subsection{Validation on Laser Scans}
In the first experiment we reproduced the results presented on the laser-scan Paris-rue-Madame dataset \cite{serna_paris-rue-madame_2014}.
The training and test data sets are generated the same way as in \cite{weinmann_distinctive_2015} and \cite{hackel_fast_2016} by randomly sampling without replacement 1000 points per each class for training, and using the rest of the points for testing.
When training a RF we achieved overall accuracies of $95.76 \%$ compared to the reported $95.75 \%$ in the 
paper although our per-class results differed slightly.
We also observed that this evaluation procedure typically yields overly optimistic accuracies, which are much higher 
than the expected accuracy on unseen test data. We noticed that such evaluation resembles an inpainting problem: given 
a few known labeled points in the cloud, estimate the labels of the rest that lie in-between. This gives a bias to the 
results and does not represent the classifier's ability to generalize to unseen datasets.

To overcome these issues we propose to split the data set into two non-overlapping regions, train on one half and test 
on the other, as described in Sec.~\ref{sec:experimentalsetup}.
If the Paris-rue-Madame dataset is split this way our overall accuracy is reduced to $\sim 90 
\%$. We believe this is a less biased estimator of the performance on unseen data, and adopt this strategy to evaluate 
performance in the rest of our experiments.

It is worth noting that the Paris-rue-Madame dataset contains only small quantities of some classes such as 
vegetation and human made objects which were found to be harder to classify correctly by \cite{hackel_fast_2016}.

\subsection{Experiments on Aerial Photogrammetry Data}

\paragraph{Intra-dataset results}
\label{sec:results-intratest}
\begin{figure*}[htp]
	\centering
	\includegraphics[width=0.92\linewidth]{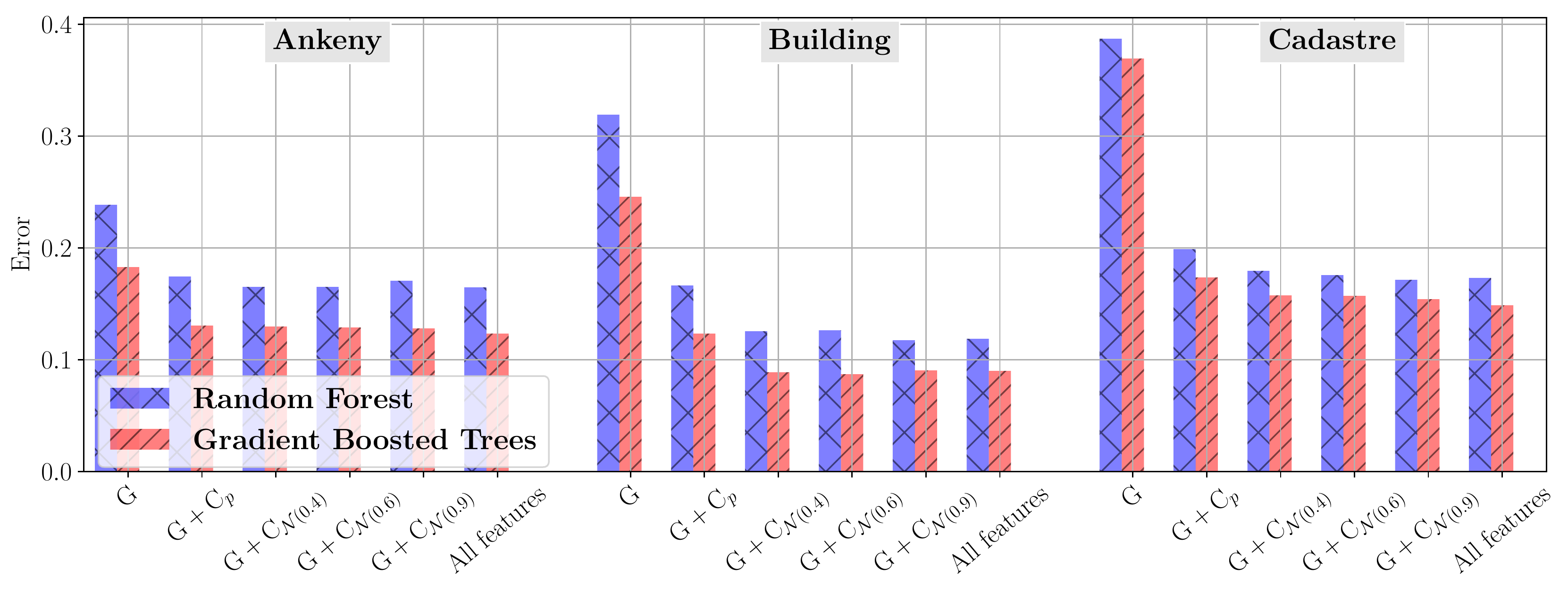}
	\caption{Intra-dataset results: classification error (number of misclassified points divided by overall number of points, the lower the better) when training and testing on different parts of the same 
	dataset. Each dataset is split into physically disjoint training and testing sets through a vertical plane.
		Incorporating color features $\mathcal{C}_{\mathcal{N}(0.6)}$ yields a significant improvement.
		The best results are obtained when combining both geometric and color features. For a detailed discussion see section \ref{sec:results-intratest}.}
	\label{fig:abserr-split}
\end{figure*}

The misclassification errors for different sets of features are presented in Fig.~\ref{fig:abserr-split}, where we can 
see that color features bring a significant improvement. The best results are obtained with 
the $\mathcal{C}_{\mathcal{N}(0.6)}$ features or with \emph{All features}. A 
second important observation is that GBT consistently outperforms RF, in some cases by a large margin.

\begin{table}[tbp]
	\centering
	\footnotesize
	\begin{tabular}{llccc}
		\toprule
		{Section} & & {Ankeny} & {Building} & {Cadastre} \\
		\midrule
		Geom. features && 67s & 23s & 46s\\
		Geom. + Color $\:\mathcal{C}_{\mathcal{N}(0.6)}$ && 105s & 56s & 53s\\
		\midrule
		\multirow{2}{*}{Random Forest (RF)} & Train & 74s & 76s & 77s \\
		& Predict & 45s & 18s & 25s \\
		\midrule
		\multirow{2}{*}{Boosted Trees (GBT)} & Train & 5s & 5s & 5s \\
		& Predict & 56s & 23s & 41s \\
		\bottomrule
	\end{tabular}
	\caption{Timings for feature computation, classifier training and prediction.
		Our whole pipeline runs in
		less than 3 minutes with any of the provided point clouds, being suitable for interactive applications.}
	\label{tab:timings}
\end{table}

\paragraph{Inter-dataset results}
\label{sec:results-crosstest}

\begin{figure*}[htp]
	\centering
	\includegraphics[width=0.92\linewidth]{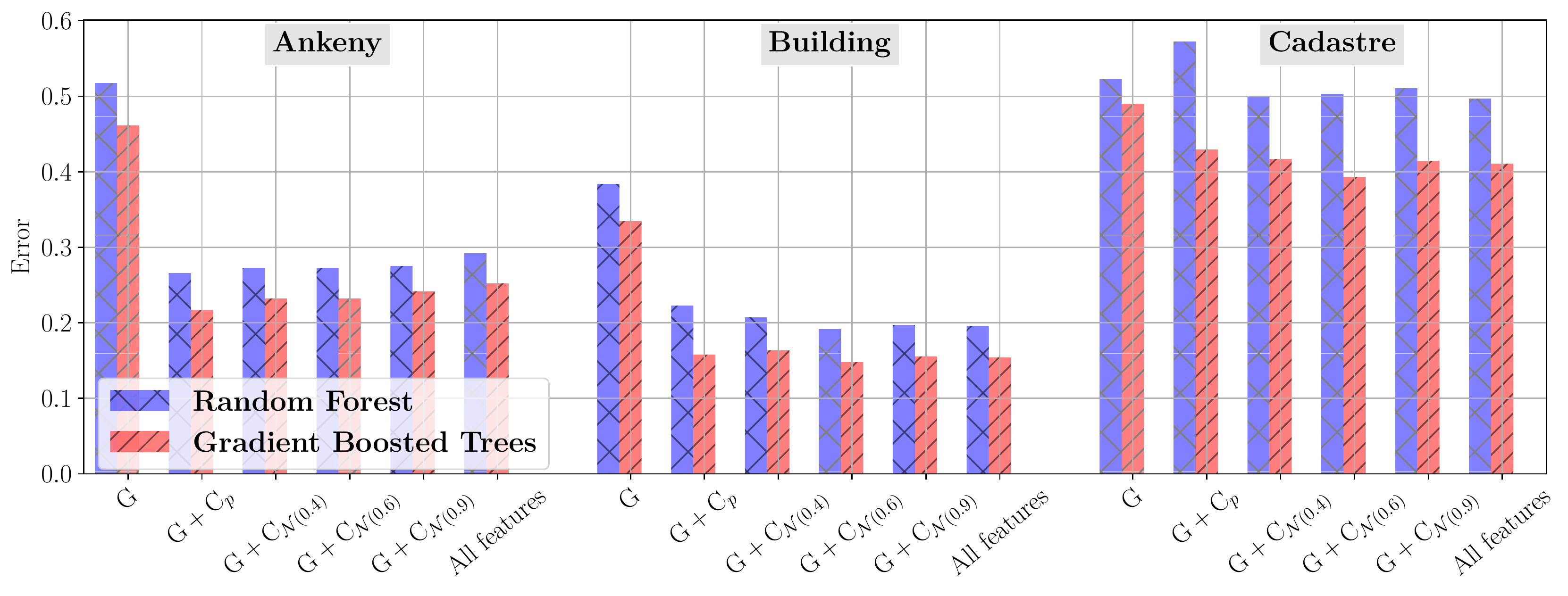}
	\caption{Inter-dataset results: classification error (number of misclassified points divided by overall number of points, the lower the better) when training on two datasets and 
	testing on the remaining one. 
		Similar to Fig.~\ref{fig:abserr-split}, the best results are obtained with both geometric and
		color features $\mathcal{C}_{\mathcal{N}(0.6)}$.
		The results on Ankeny and Cadastre show higher classification errors than those obtained when training and 
		testing on the same dataset in Fig.~\ref{fig:abserr-split}. This is due to the lack of training data for 
		certain objects or terrain type. For a detailed discussion see section \ref{sec:results-crosstest}.}
	\label{fig:abserr-crosstest}
\end{figure*}

\begin{table*}[tbp]
	\centering
	\footnotesize
	\begin{tabular}{cclcccccc@{\extracolsep{0.5cm}}c}
		\toprule
		\multirow{3}{*}{Training Datasets} &
		\multirow{3}{*}{Testing Dataset} & \multirow{3}{*}{True Label} & \multicolumn{6}{c}{Predicted Label} & \multirow{3}{*}{Overall 
			error}\\
		\cmidrule(l){4-9}
		& & &  \multirow{2}{*}{Ground} & High & 
		\multirow{2}{*}{Building} & 
		\multirow{2}{*}{Road} & \multirow{2}{*}{Car} & Human- & \\
		& & & & vegetation & & & & made object & \\
		
		\midrule
		\multirow{2}{*}{\textbf{Building}} &\multirow{3}{*}{\textbf{ANKENY}} & Ground & 52\% & 3.3\% & 0.1\% & \textbf{12\%} & 0.4\% & 0.7\% & 16.5\%\\
		\multirow{2}{*}{\textbf{Cadastre}}& & High vegetation & \textbf{3.4\%} & 7.4\% & 0.1\% & 0.2\% & 0.2\% & 0.1\% & 4.0\%\\
		& & Human-made obj. & \textbf{0.4\%} & 0.3\% & 0.1\% & 0.1\% & 0.2\% & 0.1\% & 1.1\%\\
		
		\midrule
		\multirow{2}{*}{\textbf{Ankeny}} &\multirow{3}{*}{\textbf{BUILDING}} & Building & 0.1\% & \textbf{2.6\%} & 29\% & 0.3\% & 0.9\% & 0.9\% & 4.8\%\\
		\multirow{2}{*}{\textbf{Cadastre}} & & Road & \textbf{2.2\%} & 0.3\% & 0.02\% & 33\% & 0.3\% & 0.3\% & 3.1\%\\
		& & Ground & 7.3\% & \textbf{2.1\%} & 0.01\% & 0.8\% & 0.06\% & 0.2\% & 3.1\%\\
		
		\midrule
		\multirow{2}{*}{\textbf{Ankeny}} &\multirow{3}{*}{\textbf{CADASTRE}} & Ground & 32\% & 2.1\% & \textbf{6.6\%} & 4.6\% & 0.9\% & 4.7\% & 
		18.9\%\\
		\multirow{2}{*}{\textbf{Building}} & & High vegetation & 0.6\% & 2.5\% & \textbf{7.7\%} & 0.08\% & 0.5\% & 1.4\% & 10.3\%\\
		& & Road & \textbf{2.9\%} & 0.05\% & 0.9\% & 13\% & 2.1\% & 1.6\% & 7.5\%\\
		\bottomrule
	\end{tabular}
	
	\caption{Confusion matrix for the top-3 misclassified classes. In bold we highlighted the misclassification error corresponding to the class with which the true label is confused the most. Results obtained for the $\mathcal{G} + 
		\mathcal{C}_{\mathcal{N}(0.6)}$ features with the GBT classifier, training on two datasets and testing on the 
		remaining third one. Percentages are with respect to the total number of points in the testing dataset.
		See section~\ref{sec:results-crosstest} for a detailed analysis.}
	\label{tab:prediction-breakdown}
\end{table*}

The results are shown in Fig.~\ref{fig:abserr-crosstest}. First, there is an overall increase of classification error, 
in particular for the Cadastre dataset. To analyze the results in more detail we computed the confusion matrix for the 
top-3 classes that contribute to the misclassification error, as shown in Table~\ref{tab:prediction-breakdown}. We now 
discuss the result of each dataset in detail.

\paragraph{Ankeny} The classifier performs very well for buildings and roads, as shown in 
Fig.~\ref{fig:ankeny-qualit}(b). 
However it confuses large amounts of ground points as roads. This is not surprising since most of such mistakes occur 
in croplands, which are not present in any other dataset. Finally, although high vegetation appears in the top-3 
misclassified classes in Table~\ref{tab:prediction-breakdown}, this is mostly due to ambiguities in the ground truth: 
some bushes were manually labeled as ground, while the classifier predicts them as high vegetation. 

\paragraph{Building} The classifier performs very well on this dataset. 
The highest error is due to predicting buildings as high vegetation or human-made objects.
This dataset has the lowest GSD (or highest resolution), and facades of the buildings are well-reconstructed.
This is not the case for the other two datasets with higher GSD, where few facade points are available. 
We hypothesize that the classifier is confused with the facades, finding the vegetation or human-made object to be the 
closest match.

\paragraph{Cadastre} The classifier predicts vast amounts of ground and 
vegetation points as buildings and human-made objects, leading to a very high error rate. 
This result is expected considering Table~\ref{tab:datasets-cnt}, as the Cadastre dataset contains hills and non-flat 
ground surfaces, which are not present in any of the other two datasets. 
Thus, the classifier confuses points in the regions of inclined ground with other classes that are closer in feature 
space to the training data (e.g. building roofs present a slope that resembles the properties of the points on a hill).

The analyses above highlight the importance of reliable and varied training data, in that it should resemble the unseen 
data on which the classifier will be applied, e.g different landscapes, seasons, shapes of buildings, etc.

\subsection{Qualitative Results}

\newcolumntype{R}[1]{>{\raggedright\arraybackslash}m{#1}}
\begin{figure}[t!]
	\centering
	{\large Ankeny dataset} \vspace{2em}\\
	\includegraphics[height=0.35\linewidth]{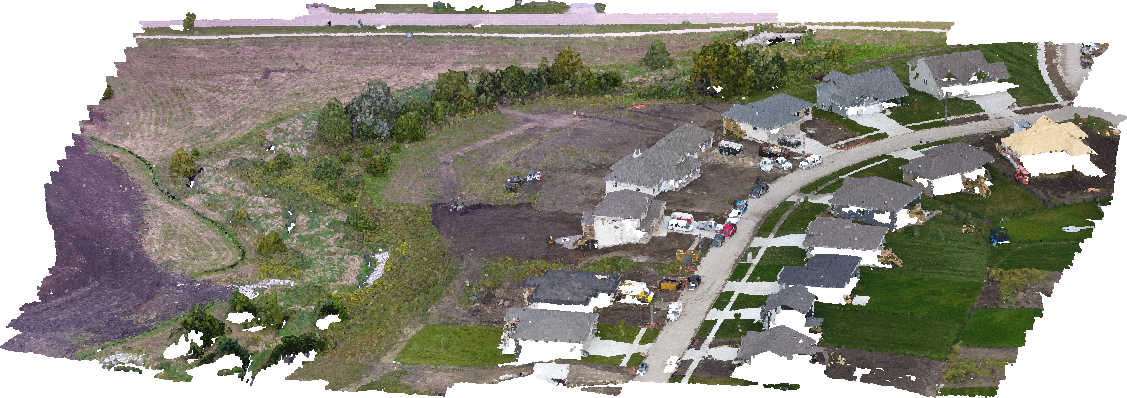}\\
	\vspace{0.4em}(a) Original data. \vspace{2em}\\
	\includegraphics[height=0.35\linewidth]{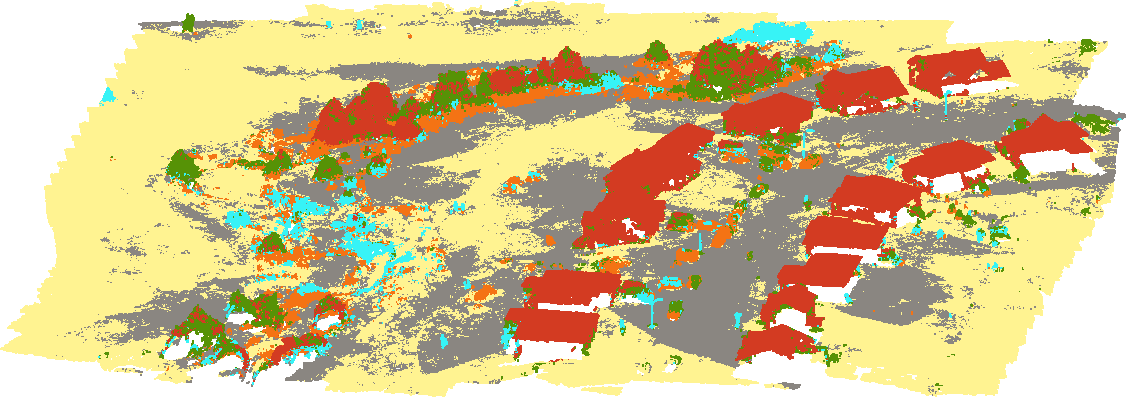}\\
	\vspace{0.4em}(b) Classification with geometry features only. \vspace{2em}\\
	\includegraphics[height=0.35\linewidth]{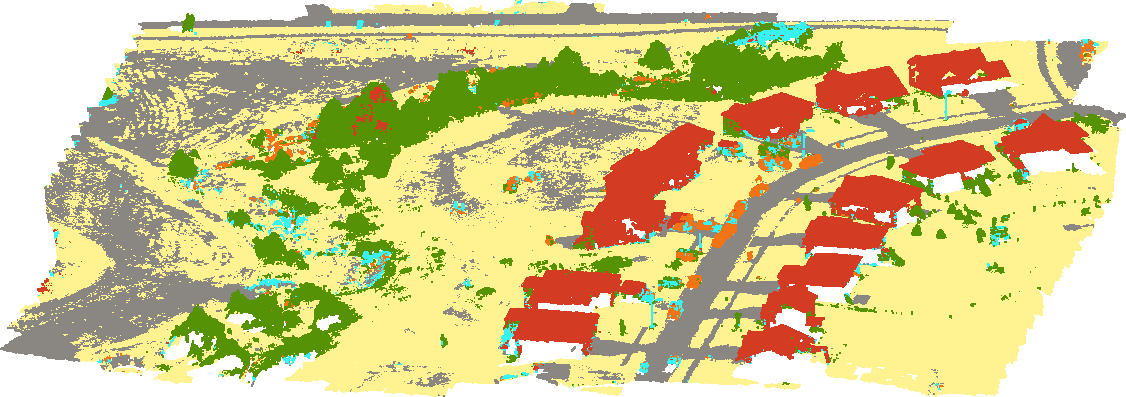}\\
	\vspace{0.4em}(c) Classification with geometry + color features.\\
	
	\vspace{2.5em}
	\setlength{\tabcolsep}{0.3em} 
	\begin{tabular}{R{0.4cm} R{1.5cm} R{0.4cm} R{1.2cm} R{0.4cm} R{2.3cm}}
		\crule[FFF391]{0.4cm}{0.4cm} & Ground & \crule[888A85]{0.4cm}{0.4cm} & Road &
		\crule[EF2929]{0.4cm}{0.4cm} & Building\\
		\crule[4E9A06]{0.4cm}{0.4cm} & High veg. & \crule[F57900]{0.4cm}{0.4cm} & Car & \crule[3FF3F6]{0.4cm}{0.4cm} & 
		Human-made obj.\\
	\end{tabular}

	\caption{Qualitative results obtained with our approach on the Ankeny dataset, using the other two datasets for 
		training. We used neighbor color features within a 0.6-meter radius neighborhood and the Gradient Boosted Trees 
		classifier. Incorporating color information into the classifier yields a significant boost in accuracy, 
		particularly for the road and high vegetation classes.}
	\label{fig:ankeny-qualit}
\end{figure}

\begin{figure}[t!]
	\centering
	{\large Buildings dataset} \vspace{2em}\\
	\includegraphics[height=0.35\linewidth]{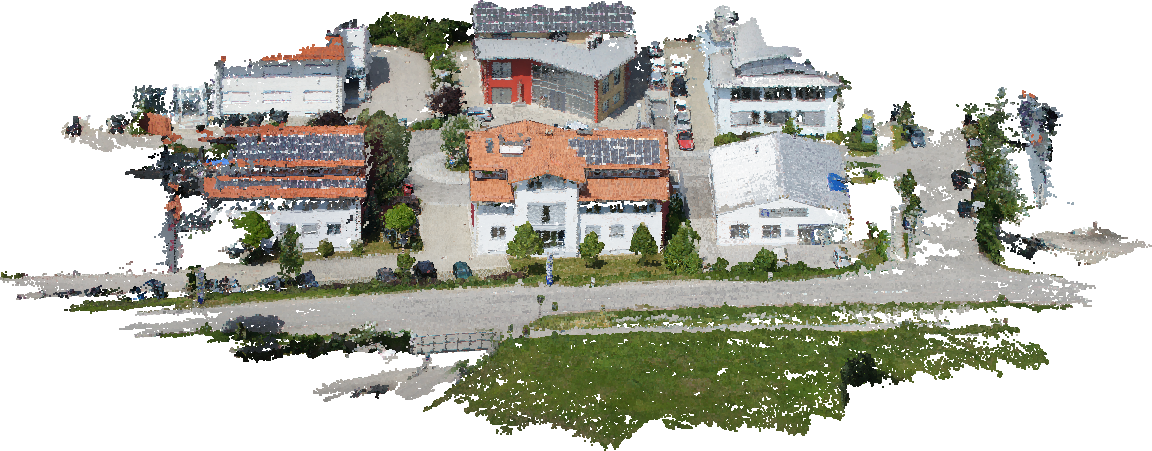}\\
	\vspace{0.4em}(a) Original data. \vspace{2em}\\
	\includegraphics[height=0.35\linewidth]{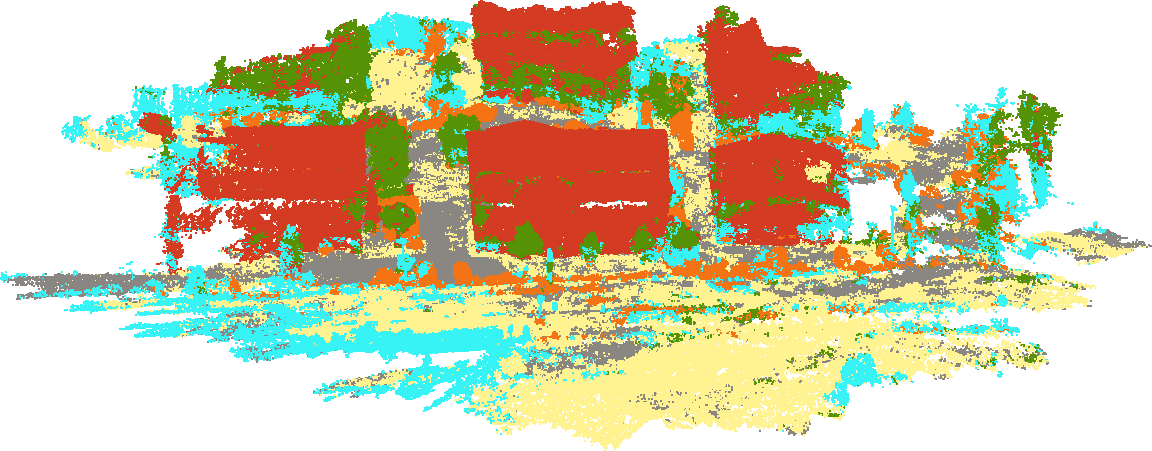}\\
	\vspace{0.4em}(b) Classification with geometry features only. \vspace{2em}\\
	\includegraphics[height=0.35\linewidth]{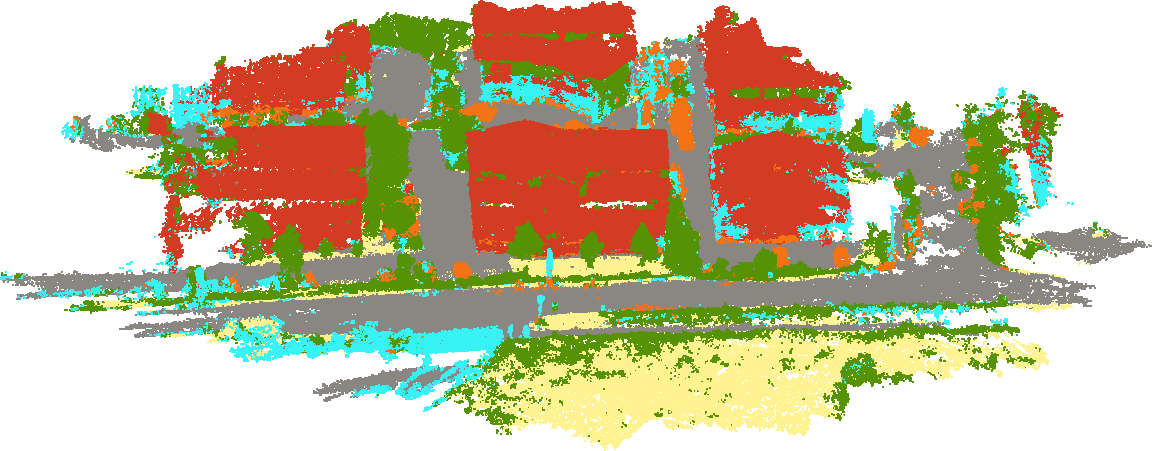}\\
	\vspace{0.4em}(c) Classification with geometry + color features.\\
	
	\vspace{2.5em}
	\setlength{\tabcolsep}{0.3em} 
	\begin{tabular}{R{0.4cm} R{1.5cm} R{0.4cm} R{1.2cm} R{0.4cm} R{2.3cm}}
		\crule[FFF391]{0.4cm}{0.4cm} & Ground & \crule[888A85]{0.4cm}{0.4cm} & Road &
		\crule[EF2929]{0.4cm}{0.4cm} & Building\\
		\crule[4E9A06]{0.4cm}{0.4cm} & High veg. & \crule[F57900]{0.4cm}{0.4cm} & Car & \crule[3FF3F6]{0.4cm}{0.4cm} & 
		Human-made obj.\\
	\end{tabular}
	
	\caption{Qualitative results obtained with our approach on the Buildings dataset, using the other two datasets for 
		training. We used neighbor color features within a 0.6-meter radius neighborhood and the Gradient Boosted Trees 
		classifier. Incorporating color information into the classifier results improves classification,
		particularly for the roads between buildings.}
	\label{fig:buildings-qualit}
\end{figure}

\begin{figure}[ht]
	\centering
	{\large Cadastre dataset} \vspace{2em}\\
	\includegraphics[width=0.98\linewidth]{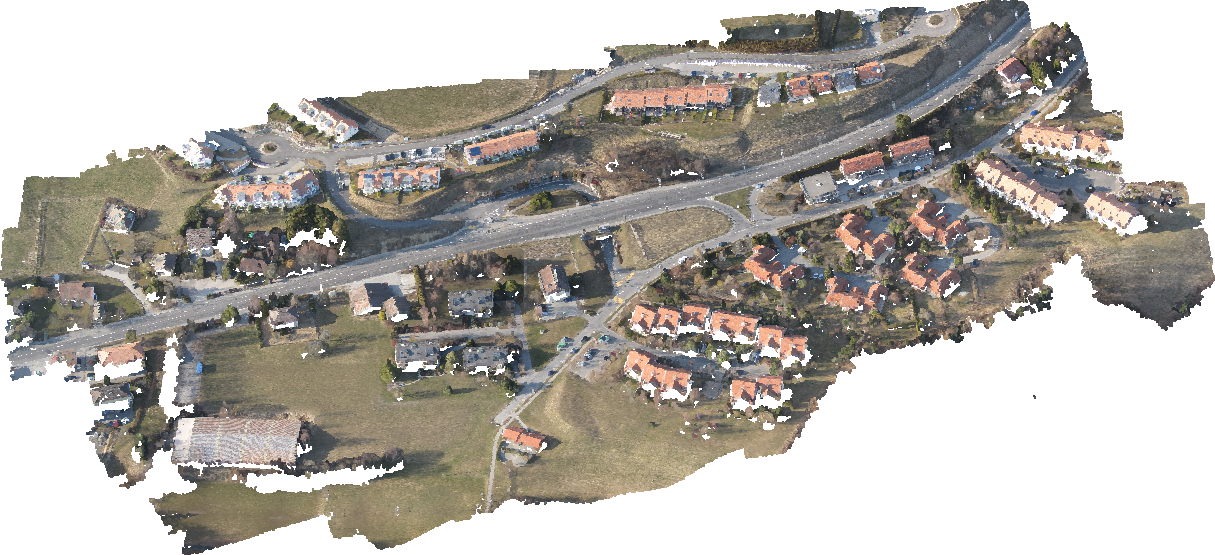}\\
	\vspace{0.4em}(a) Original data. \vspace{2em}\\
	\includegraphics[width=0.98\linewidth]{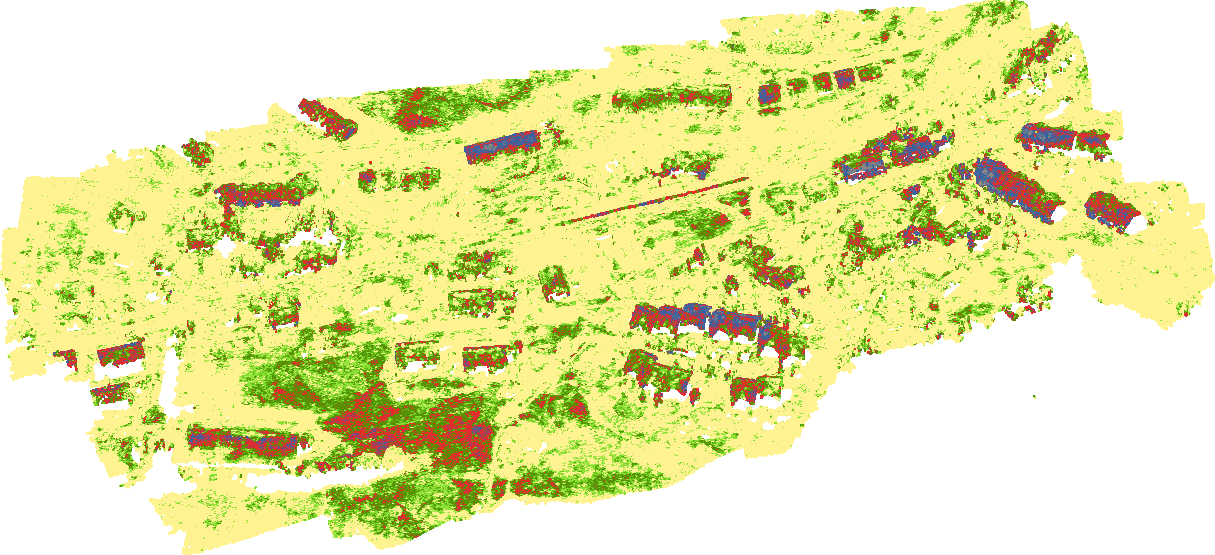}\\
	\vspace{0.4em}(b) Classification with geometry features only. \vspace{2em}\\
	\includegraphics[width=0.98\linewidth]{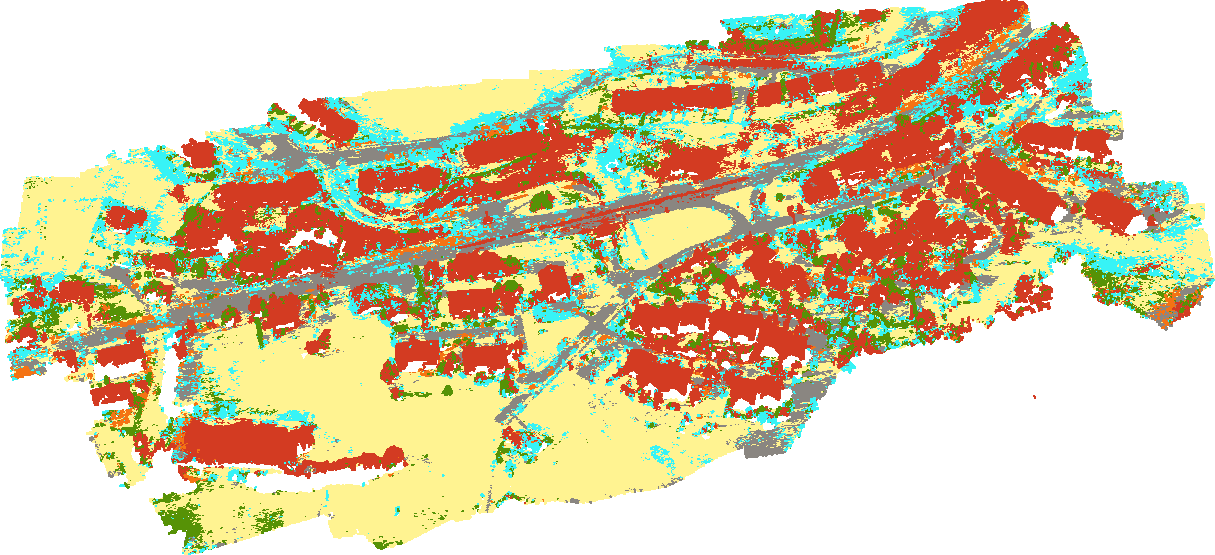}\\
	\vspace{0.4em}(c) Classification with geometry + color features.\\
	
	\vspace{2.5em}
	\setlength{\tabcolsep}{0.3em} 
	\begin{tabular}{R{0.4cm} R{1.5cm} R{0.4cm} R{1.2cm} R{0.4cm} R{2.3cm}}
		\crule[FFF391]{0.4cm}{0.4cm} & Ground & \crule[888A85]{0.4cm}{0.4cm} & Road &
		\crule[EF2929]{0.4cm}{0.4cm} & Building\\
		\crule[4E9A06]{0.4cm}{0.4cm} & High veg. & \crule[F57900]{0.4cm}{0.4cm} & Car & \crule[3FF3F6]{0.4cm}{0.4cm} & 
		Human-made obj.\\
	\end{tabular}
	
	\caption{Qualitative results obtained with our approach on the Cadastre dataset, using the other two datasets for 
		training. We used neighbor color features within a 0.6-meter radius neighborhood and the Gradient Boosted Trees 
		classifier. Using color information results in a more accurate segmentation.}
	\label{fig:cadastre-qualit}
\end{figure}

Figures~\ref{fig:ankeny-qualit}, \ref{fig:buildings-qualit} and \ref{fig:cadastre-qualit} show 3D views of each dataset 
and the respective classified point clouds obtained when using geometric features only, as well as with our approach. 
Overall the results are very satisfying, especially when one considers the heterogeneity of the different datasets, as 
discussed earlier. These figures also help clarify the observations described in 
section~\ref{sec:results-crosstest}. For example, grass and ground on slopes in Cadastre are sometimes misclassified as 
roof in Fig.~\ref{fig:cadastre-qualit}(c).

\subsection{Timings}

Table~\ref{tab:timings} shows the break down of the timings obtained on a 6-core Intel i7 
3.4~GHz computer. Our approach is very efficient, taking less than 3 minutes to classify 
every point in any of the presented photogrammetry datasets. This makes it ideal for the applications where the user needs to interact with the software to correct the 
training data or fix the classifier's predictions.

\begin{table}[tbp]
	\centering
	\footnotesize
	\begin{tabular}{llccc}
		\toprule
		\textbf{Testing dataset} & & \textbf{Ankeny} & \textbf{Building} & \textbf{Cadastre} \\
		\midrule
		Geom. features && 46\% & 33\% & 48\%\\
		\midrule
		
		Geom. + Color $\mathcal{C}_{\mathcal{N}(0.6)}$ && 24\% & 16\% & 41\%\\
		
		\bottomrule
	\end{tabular}
	\caption{Overall classification errors for inter-dataset experiments with LGBM classifier run with 2 sets of 
	features: geometric only and both geometric and color. Color information yields a significant performance boost in 
	prediction accuracy in all datasets.}
	\label{tab:condensed}
\end{table}

\section{Conclusion}
\label{sec:conclusion}

In this paper we described an approach for a point-wise semantic labeling of aerial photogrammetry point clouds.
The core contribution of our work is the use of color features, what improves significantly the overall classification results.
Further we provide a concise comparison between two standard machine learning techniques that hopefully facilitates the 
decision making process of future research, showing that the Gradient Boosted Trees classifier outperforms the Random 
Forest classifier, in some cases by a large margin.
Our method performs not only with high accuracies over the whole range of datasets used in the experiments but also 
with a high computational efficiency, making our approach suitable for interactive applications.

There are several directions that we would like to explore to increase accuracy and further decrease computational complexity. First we would like to explore the possibility to use \textit{auto-context} information to train a second classifier that takes into account class labels of the neighboring points.
Our preliminary experiments with this technique provided smoother results and higher accuracy.
Another interesting topic to is to combine point cloud and image classification.
Existing image classification algorithms are trained specifically to detect objects of such classes as cars and other human made objects. 
This method could increase the robustness of our classifier in particular on the classes that often confused: cars and 
human-made objects.

The classification method presented in this paper will soon be part of Pix4Dmapper Pro.
Earlier we mentioned that access to properly labeled training data that represents aerial 
photogrammetry point clouds is limited. To overcome this issue we will implement an incremental training method, where 
users will be given the possibility to classify their data, visualize and correct errors manually.
In a next step we will offer our users the possibility to include their datasets into our training data to improve the 
classifier quality. As the amount of training data increases we will be able not only to provide more accurate 
classifiers but to also train specialized ones. For example, we could provide a selection of classifiers for indoor and 
outdoor scenes, and for different seasons and scales.

{
	\begin{spacing}{0.9}
		\bibliography{bibliography} 
	\end{spacing}
}

\end{document}